\pdfoutput=1

\documentclass[11pt]{article}

\usepackage[preprint]{acl}

\usepackage{times}
\usepackage{latexsym}

\usepackage{CJKutf8}
\usepackage{amsmath}
\usepackage{amsthm}
\usepackage{amsfonts}
\usepackage{makecell}
\usepackage[caption=false,font=normalsize,labelfont=sf,textfont=sf]{subfig}
\usepackage{multirow}
\usepackage{anyfontsize}

\usepackage[pdftex]{graphicx}
\usepackage{algorithm,algpseudocode}
\usepackage{booktabs}

\newtheorem{theorem}{Theorem}[section]

\newtheorem{definition}{Definition}

\usepackage[T1]{fontenc}

\usepackage[utf8]{inputenc}

\usepackage{microtype}

\usepackage{inconsolata}

\usepackage{graphicx}
\usepackage{xcolor,pifont}

\newcommand*\colourcheck[1]{%
  \expandafter\newcommand\csname #1check\endcsname{\textcolor{#1}{\ding{52}}}%
}

\colourcheck{green}

\newcommand*\colourcross[1]{%
  \expandafter\newcommand\csname #1cross\endcsname{\textcolor{#1}{\ding{56}}}%
}

\colourcross{red}

\title{Augmenting Compliance-Guaranteed Customer Service Chatbots: Context-Aware Knowledge Expansion with Large Language Models}

\author{Mengze Hong$^{1, 2}$, Chen Jason Zhang$^{1}$, Di Jiang$^{1}$\thanks{Corresponding Author}, Yuanqin He$^{2}$\\ 
$^{1}$Hong Kong Polytechnic University, $^{2}$AI Group, WeBank Co., Ltd \\
}

\begin{document}
\maketitle
\begin{abstract}
Retrieval-based chatbots leverage human-verified Q\&A knowledge to deliver accurate, verifiable responses, making them ideal for customer-centric applications where compliance with regulatory and operational standards is critical. To effectively handle diverse customer inquiries, augmenting the knowledge base with ``similar questions'' that retain semantic meaning while incorporating varied expressions is a cost-effective strategy. In this paper, we introduce the Similar Question Generation (SQG) task for LLM training and inference, proposing context-aware approaches to enable comprehensive semantic exploration and enhanced alignment with source question-answer relationships. We formulate optimization techniques for constructing in-context prompts and selecting an optimal subset of similar questions to expand chatbot knowledge under budget constraints. Both quantitative and human evaluations validate the effectiveness of these methods, achieving a 92\% user satisfaction rate in a deployed chatbot system, reflecting an 18\% improvement over the unaugmented baseline. These findings highlight the practical benefits of SQG and emphasize the potential of LLMs, not as direct chatbot interfaces, but in supporting non-generative systems for hallucination-free, compliance-guaranteed applications.
\end{abstract}

\begin{figure*}[t!]
  \centering
\includegraphics[width=0.84\textwidth] {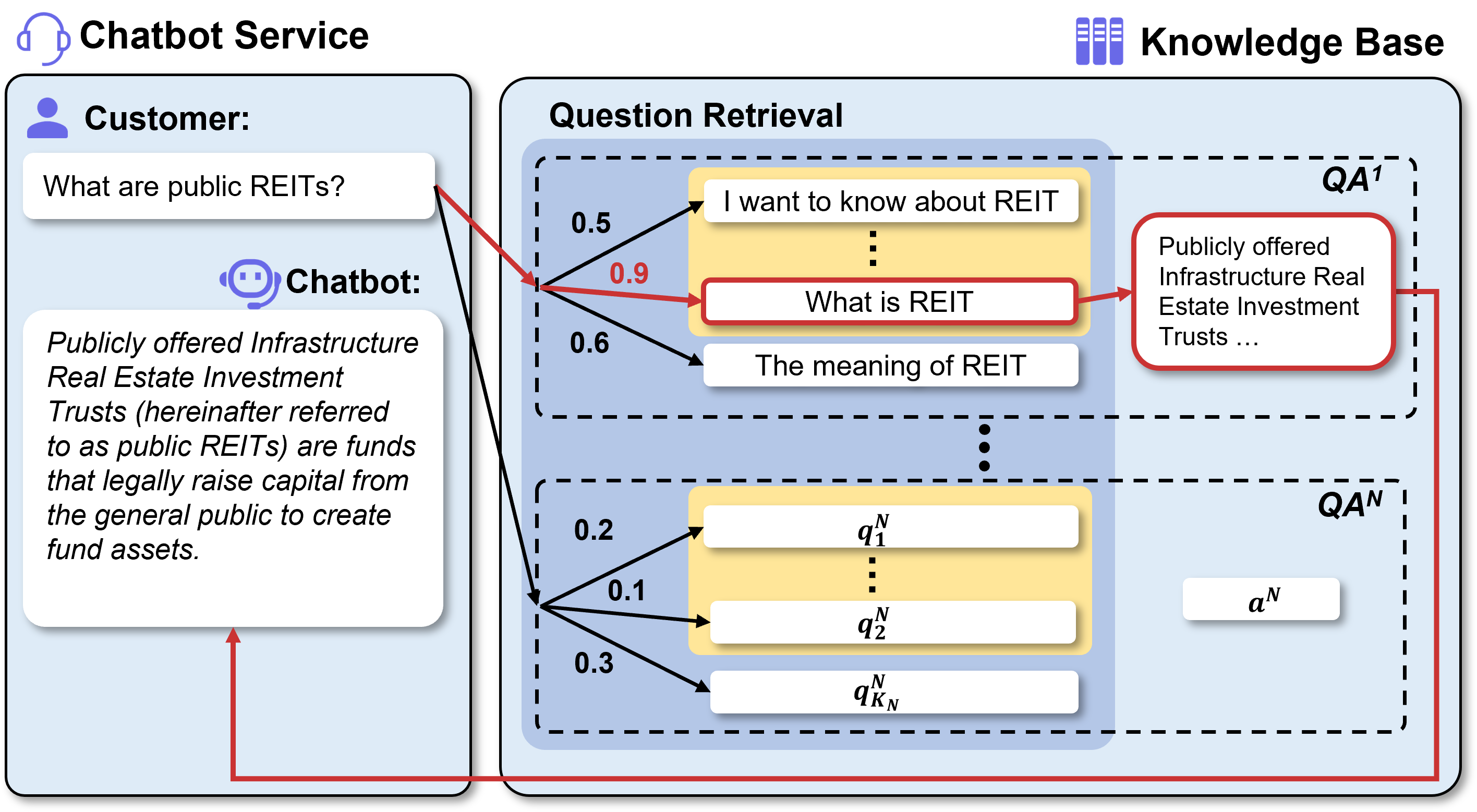}
  
  \caption{Schematic overview of a compliance-guaranteed chatbot with a predefined knowledge base for Match-and-Respond. The yellow region highlights the questions augmented by the similar question generation.} 
  \label{fig:process}
\end{figure*}

\section{Introduction}

Customer service automation is essential for digital transformations, commonly deploying AI-driven chatbots to handle diverse inbound customer inquiries to reduce the workload in labor-intensive call centers and enable timely responses across online platforms \cite{jiang2025chatmap}. However, popular generative language models are prone to hallucinations, generating inconsistent or incorrect answers \cite{huang2025survey}, making LLM-driven chatbot interfaces infeasible in sectors like healthcare and finance, where reliability and verifiability of responses are critical concerns \cite{8509382, 10.14778/3137765.3137820}. To ensure compliance with regulatory and operational requirements, retrieval-based chatbot systems, a framework established before LLMs, use human-verified question-answer (QA) pairs from an offline knowledge base to deliver \textbf{hallucination-free responses} \cite{wu2018response}. As illustrated in Figure \ref{fig:process}, these systems employ a Match-and-Respond process, matching input queries to existing questions to retrieve accurate responses, eliminating generation needs.

\begin{CJK}{UTF8}{gbsn}

\begin{table}[!t]
    \centering
    \resizebox{\columnwidth}{!}{
    \begin{tabular}{c|l}
        \toprule
        Customer & \thead[l]{我开了证明怎么还没收到？\\  I applied for the certificate, why haven’t I received it yet?} \\ \midrule
        Source   & \thead[l]{证明开具时间要多久？\\How long does it take to process the certificate? } \redcross \\ \midrule
        Generated  & \thead[l]{开了证明什么时候能收到？\\Applied for the certificate, when can I receive? } \greencheck \\ 
        \bottomrule
        \end{tabular}}
    \caption{Customer query matching with source question (failed) and generated similar question (success).}
    \label{tab:demo}
\end{table}

\end{CJK}

In practice, customer queries exhibit high diversity in expression, and a failure in query matching may interrupt the interaction and cause user dissatisfaction \cite{zhang-etal-2025-fine, wang-etal-2024-blendfilter}. Expanding chatbot knowledge offline enables seamless integration into production systems, offering a static augmentation approach to enhance query-matching performance with minimal computational load and runtime latency, ensuring practicality for lightweight industrial deployment. This expansion can be effectively achieved through the \textbf{Similar Question Generation (SQG)} task, where each source question in the predefined knowledge base is augmented with multiple ``similar questions'' that are diverse in expression while preserving semantic consistency to maintain the question-answer relationship. As demonstrated in Table \ref{tab:demo}, a similar question more effectively matches the user query.

Traditionally, data augmentation relies on costly human crowdsourcing, offering limited diversity due to its independent nature \cite{liu-etal-2022-toward}. Rule-based automation methods, such as SimBERT \cite{simbert} and RoFORMER-Sim \cite{roformersim}, improve semantic consistency but lack contextual awareness, producing repetitive or generic outcomes \cite{feng2021survey}. The emergence of LLMs highlights their potential for data augmentation through advanced language understanding and generation capabilities \cite{wei2022emergent, hong2025llm}, improved by prompting \cite{liu2023pre} and fine-tuning \cite{bao2023tallrec} techniques, which frequently outperform human workers in cost efficiency and performance \cite{gilardi2023chatgpt, tornberg2023chatgpt}. However, the SQG task presents a unique challenge in requiring diversity among generated questions, where standard sequence-to-sequence methods struggled to produce varied questions due to limited control over the generation process \cite{jiang2018sequence}, necessitating tailored model training and inference strategies.

In this work, we propose novel approaches for augmenting customer service chatbots with similar questions generated by LLMs, highlighting the importance of contextual awareness in the guided generation process. The contributions include:

\begin{enumerate}
\item To the best of our knowledge, we are the first to define the SQG task for retrieval-based service chatbot augmentation, formulating LLM training and inference, and proposing context-aware one-to-many generation paradigms.
\item We present budget-constrained optimization techniques to select prompt demonstrations and similar question subsets, facilitating knowledge base expansion and ensuring cost-effective, adaptable deployment across diverse real-world application scenarios.
\item Experiments demonstrate over 120\% relative improvement in qualitative assessment, 4.74\% increase in total diversity, and 18\% enhancement in user satisfaction compared to unaugmented chatbot systems.
\end{enumerate}

\section{Problem Formulation and Background}
\label{sec:method}

\begin{CJK}{UTF8}{gbsn}
\begin{table*}[ht]
\centering
\scriptsize
\resizebox{\textwidth}{!}{
    \begin{tabular}{{p{2.6cm}p{10cm}}}
        \toprule
        \textbf{Method} & \textbf{Prompt Template}\\ 
        \midrule
        \makecell[l]{\textbf{One-to-one Generation} \\ \textbf{(Standard)}} &  \makecell[l]{\textbf{Instruction:} 将 ``\{原问题\}'' 改写为保持相同意义但表述不同的新问句。\\ (Rewrite ``\{source question\}'' to maintain the same meaning but express it differently in a new sentence.)\\\textbf{Response:} 相似问题 (similar question)}  \\
        \midrule
        \makecell[l]{\textbf{Context-Aware} \\ \textbf{Batch Generation}} & \makecell[l]{\textbf{Instruction:} 生成K条与 ``\{原问题\}'' 不同且意思相近的问题。 \\(Generate K different yet closely related similar questions based on the question ``\{source question\}''.) \\  \textbf{Response:} \{相似问题1, \dots, 相似问题K \}\quad\{(similar question 1, \dots, similar question K)\}} \\
        \midrule
        \makecell[l]{\textbf{Intention-Enhanced} \\ \textbf{Batch Generation}} & \makecell[l]{\textbf{Instruction:} 根据问题 ``\{原问题\}'' 和答案 ``\{原答案\}''，生成K个不同且意思相近的问题。 \\(Generate K different yet closely related similar questions based on the question ``\{source question\}'' \\ and the answer ``\{source answer\}''.) \\  \textbf{Response:} \{相似问题1, \dots, 相似问题K \}\quad\{(similar question 1, \dots, similar question K)\}} \\
    \bottomrule
    \end{tabular}}
\caption{Illustration of conventional generation and proposed methods for fine-tuning and inference of LLMs.}
\label{table:prompt_template}
\end{table*}
\end{CJK}

\subsection{Problem Formulation}

Similar Question Generation aims to create a diverse yet semantically consistent set of questions that can be matched to a specific answer in a knowledge base. In this context, \textbf{semantic consistency} refers to the preservation of the original intent and meaning (e.g., ``inquire-promotion''), ensuring the generated questions can still be accurately matched to the correct answer in the knowledge base \cite{gollapalli-ng-2022-qsts, hong2025dialin, jiang2021familia}. Conversely, \textbf{syntactic diversity} pertains to the variation in phrasing and structure of the generated questions, enabling different expressions in the knowledge base that are essential for enhancing query-matching  \cite{guo-etal-2024-curious, ma2023query}.

\subsection{SQG Training and Inference with LLMs}

Conventional methods that utilize LLMs for similar question generation typically adhere to a naive sequence-to-sequence approach, referred to as the \textbf{one-to-one} paradigm. In this approach, the LLM generates a single question at a time in response to a given source question. For a set of similar questions \( (q_1, \dots, q_K) \), we can construct training samples by pairing questions, such as \(\{(q_1, q_2), \dots, (q_1, q_K)\}\). A typical prompt template is illustrated in the first block of Table \ref{table:prompt_template}. Given a generative language model \( P_{\Phi}(y|x) \) with parameters \( \Phi \), the training objective can be formulated as maximizing the following language modeling objective:

\begin{equation}
\begin{aligned}
\label{equ:goal}
    \mathcal{L}_{ft} = - & \sum_j \sum_i log(P_{\Phi}(q_{j}|q_{i})).
\end{aligned}
\end{equation}

\section{Proposed Methods}

\subsection{Context-Aware Batch Generation}
To enhance control over the generation process, we introduce the \textbf{one-to-many} paradigm. This method enables the LLM to generate multiple similar questions in response to a single source question (see the second block of Table \ref{table:prompt_template}). During training, the LLM learns to identify semantic similarities and subtle expressive differences among multiple target questions. In the inference phase, the auto-regressive nature of LLMs \cite{NIPS2017_3f5ee243} allows for the incorporation of previously generated questions, which helps regularize subsequent outputs and reduces the likelihood of generating repetitive or excessively divergent questions. While one-to-many generation, or Batch Prompting \cite{cheng-etal-2023-batch}, is typically used for cost-saving and often delivers lower performance compared to standard prompting, we argue that incorporating previously generated questions into autoregressive generation is highly effective for the SQG task, which introduces contextual guidance and leads to more diversified questions with lower generation cost.

\begin{figure}[!t]
\centering
(a){\includegraphics[width=1.0in]{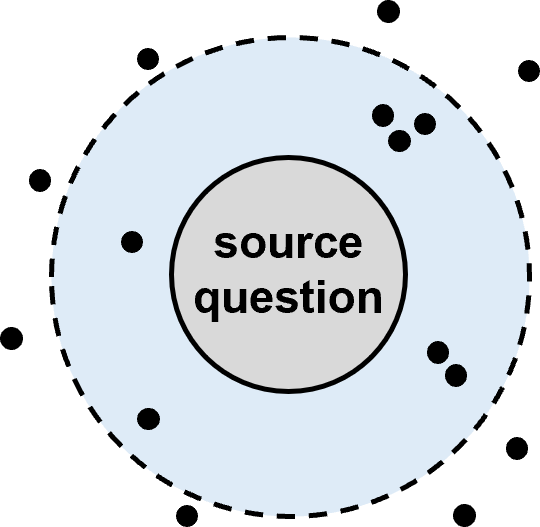}%
\label{fig_first_case}}
\hfil
(b){\includegraphics[width=1.65in]{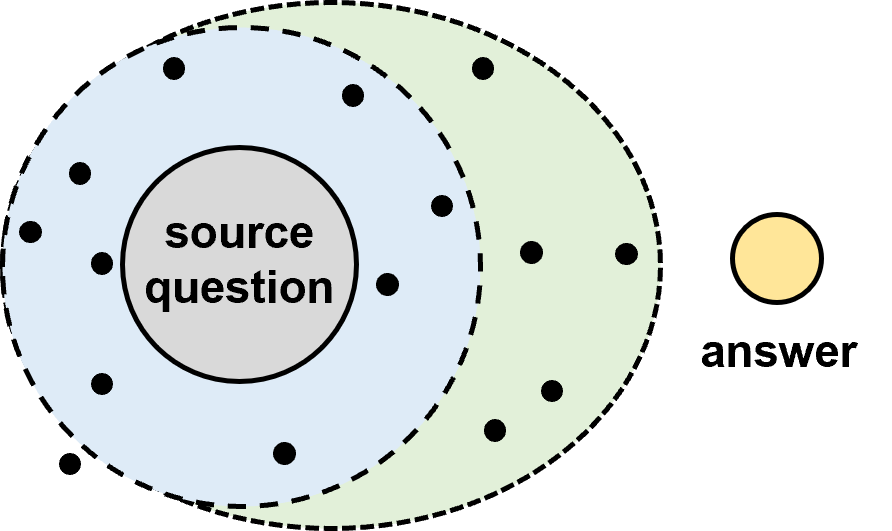}%
\label{fig_second_case}}

\caption{Illustration of the generated questions in semantic space with respect to the source question and the corresponding answer. The blue region represents the desired semantic space surrounding the source question. (a) Standard one-to-one objective: generated questions often either truncate or fall outside this desired region. (b) Intent-Enhanced Batch Generation: the green region indicates the expanded exploration region that meets the semantic consistency of the source QA pair.} 
\label{fig:semantic_space}
\end{figure}

 
\subsection{Intention-Enhanced
Batch Generation}

While the one-to-many paradigm enables a more effective exploration of the semantic space surrounding the source question, this space is constrained by the strict requirement for semantic consistency. Since the SQG is an augmentation process with known question-answer pairs, integrating the \textbf{source answer} can also be viewed as introducing contextual prior knowledge into the generation process. The corresponding formulation is presented in the third block of Table \ref{table:prompt_template}. When visualized in the semantic space, as shown in the green region of Figure \ref{fig:semantic_space} (b), this approach expands the exploration beyond the immediate vicinity of the source question and skews towards the desired answer.

\subsection{Refined Training Objective}
With the proposed one-to-many generation, we refine the training objective to generate multiple similar questions from a single QA pair. Formally, given a set of similar questions, ($q_1$, $\dots$, $q_N$), and the corresponding answer $a$, we construct training sample as (($q_i$, $a$), ($q_{j+1}$, $\dots$, $q_{j+L}$)), where $(q_i, a)$ is the input QA pair, and $(q_{j+1}$, $\dots$, $q_{j+L})$ are $L$ target questions; here, $i$ indexes the input question from the knowledge base ($i=1,\dots,M$), and $j$ iterates over starting indices of $L$ consecutive similar questions ($j=1,\dots,N-L$). The loss for predicting each target question $q_{j+l}$ is defined as:
\vspace{-0.1em}
\begin{equation*}
\resizebox{\columnwidth}{!}{$
    \begin{aligned}
        \mathcal{L}_{j+l}(q_{i}, a) = - \log(P_{\Phi}(q_{j+l} | q_{i}, a, q_{j+1}, \dots, q_{j+l-1})),
    \end{aligned}
$}
\label{eq:goal_sub}
\end{equation*}
\noindent where the generation of the question $q_{j+l}$ is conditioned on both the original QA pair $(q_i, a)$ and all previously generated similar questions in the same batch, $(q_{j+1},\dots,q_{j+l-1})$, enabling the model to capture contextual dependencies and improve syntactic diversity while maintaining semantic consistency. Consequently, the overall training objective can be formulated as:
\vspace{-0.2em}
\begin{equation}
\begin{aligned}
    \mathcal{L}_{Intention} = & \sum_i \sum_j \sum_{l=1}^L \mathcal{L}_{j+l}(q_{i},a).
\end{aligned}
\end{equation}

\section{Optimization Framework}

\begin{algorithm}[!t]
\small
\caption{Question Subset Mining}
\label{alg:qsm}
\begin{algorithmic}[1]
\Require Knowledge base \( \mathcal{D} = \{(q_i, a_i)\}_{i=1}^M \setminus \{(q_s, a_s)\} \), source question \( q_s \), target \( K \)
\Ensure Question subset \( \mathcal{P} \subseteq \mathcal{D} \)
\State \( \mathcal{P} \gets \emptyset \), \( L \gets \emptyset \)
\For{\( (q_i, a_i) \in \mathcal{D} \)}
    \State \( \phi(q_i) \gets \text{S}(q_s, q_i) \)
    \State \( L \gets L \cup \{ (q_i, a_i, \phi(q_i)) \} \)
\EndFor
\State Order \( L \) by \( \phi(q_i) \) in descending sequence
\State \( \mathcal{P} \gets \mathcal{P} \cup \{ \text{first triple in } L \} \), \( L \gets L \setminus \{ \text{first triple} \} \)
\While{\( |\mathcal{P}| < K \) and \( L \neq \emptyset \)}
    \State Select \( z' \in L \) maximizing \( \sum_{z \in \mathcal{P}} \text{dist}(q_z, q_{z'}) + \phi(q_{z'}) \)
    \State \( \mathcal{P} \gets \mathcal{P} \cup \{ z' \} \), \( L \gets L \setminus \{ z' \} \)
\EndWhile
\State \Return \( \mathcal{P} \)
\end{algorithmic}
\end{algorithm}

\subsection{Dynamic Demonstration Selection for Contextual Prompting}
To enhance contextual generation, we construct in-context prompts using the knowledge base $\mathcal{D} = {(q_i, a_i)}_{i=1}^M$, where some questions $q_i$ are associated with similar questions $\mathbf{q}_i^*$. The goal is to select $K$ examples for a target question $q_s$ and append them with their similar questions to enable in-context learning. The objective is:

\vspace{-0.5em}
\begin{equation}
\small
\resizebox{\columnwidth}{!}{$
\begin{aligned}
\arg\max_{\mathcal{P} \subseteq \mathcal{D}, |\mathcal{P}|=K} \
&\left[ \sum_{i=1}^K \text{S}(q_s, q_{p_i}) + \frac{\alpha}{K} \sum_{i \neq j} \text{dist}(q_{p_i}, q_{p_j}) \right],
\end{aligned}$}
\label{eq:optimization}
\end{equation}

\noindent where $\text{S}(q_s, q_{p_i})$ is the cosine similarity between BERT embeddings of $q_s$ and $q_{p_i}$, ensuring relevance, $\text{dist}(q_{p_i}, q_{p_j})$ is the Euclidean distance between question embeddings, measuring diversity \cite{qian2004similarity}, and $\alpha$ is a tunable constant that normalizes the diversity contribution to approximately linear scaling ($\alpha = 0.5$). To solve this optimization problem, we propose the Question Subset Mining (QSM) algorithm, designed to balance relevance and diversity (see Algorithm \ref{alg:qsm}).

\begin{algorithm}[!t]
\small
\caption{Greedy Algorithm for Maximizing Pairwise Diversity}
\label{alg:greedy}
\begin{algorithmic}[1]
\Require Set of candidates \( Q^{*} \), budget \( B \), cost function \( \text{cost}(q) \), distance function \( \text{dist}(\cdot, \cdot) \)
\Ensure Subset \( S \subseteq Q^{*} \)
\State \( S \gets \emptyset \), \( B_r \gets B \)
\While{\( B_r > 0 \) and \( Q^{*} \setminus S \neq \emptyset \)}
    \State Select \( q^* \in Q^{*} \setminus S \) that maximizes:
    \vspace{-0.8em}
    \[
    \frac{\Delta f(S, q^*)}{\text{cost}(q^*)} = \frac{\sum_{q \in S} \text{dist}(q, q^*)}{\text{cost}(q^*)}
    \]
    \vspace{-0.8em}
    \If{\( \text{cost}(q^*) \leq B_r \)}
        \State \( S \gets S \cup \{ q^* \} \)
        \State \( B_r \gets B_r - \text{cost}(q^*) \)
    \Else
        \State \( Q^{*} \gets Q^{*} \setminus \{ q^* \} \)
    \EndIf
\EndWhile
\State \Return \( S \)
\end{algorithmic}
\end{algorithm}

\subsection{Similar Question Selection for Knowledge Base Expansion}
\label{sec:selection}

Instead of generating a small set of questions directly, we argue that a two-step approach, which generates many candidate questions and then selects the best subset, would offer greater flexibility and achieve better results in industrial applications. The constrained optimization problem is formally defined as:

\vspace{-1em}
\begin{equation}
\small
\begin{aligned}
& \max_{S \subseteq Q^{*}} \sum_{\substack{q_a, q_b \in S \\ q_a \neq q_b}} \text{dist}(q_a, q_b) \quad \text{s.t.} \quad \sum_{q \in S} \text{cost}(q) \leq B.
\end{aligned}
\end{equation}
where \(\text{cost}(q)\) denotes the cost of including question \(q\), either by token length or a uniform cost.

With the proof of NP-hardness and submodularity presented in Appendix \ref{sec:optimization}, we propose a greedy algorithm that efficiently approximates the optimal solution with a guaranteed approximation bound of \( 1 - 1/e \). The greedy algorithm iteratively selects the sample \( q^* \) that provides the highest marginal gain relative to its cost while satisfying the budget constraint (see Algorithm \ref{alg:greedy}).

\begin{table*}[th!]
\centering
\resizebox{\textwidth}{!}{
\begin{tabular}{l|ccc|ccc|c}
 \hline
 \multirow{2}{*}{Models} & \multicolumn{3}{c|}{\textbf{Semantic relevance}} & \multicolumn{3}{c|}{\textbf{Character-level diversity}} & \textbf{Acceptance ratio}\\
 & Precision & Recall & F1-Score & Distinct-1 & Distinct-2 & Distinct-Avg & \\
 \hline
 SimBERT  & \underline{0.8622} & 0.7744 & 0.8160 &  0.1387 & 0.2386 & 0.1562 & $18.3\%$ \\
 RoFormer-Sim  & 0.8574 & 0.7704 & 0.8115 & 0.1836 & 0.3092 & 0.2073 & $20.0\%$ \\
 ChatGLM2 (Zero-Shot) & 0.6804 & 0.7152 & 0.6973 & \underline{\textbf{0.2607}} & \underline{\textbf{0.3889}} & \underline{\textbf{0.3248}} & -\\
 ChatGLM2 (Few-Shot)  & 0.5475 & 0.5882 & 0.5671 & 0.1752 & 0.2005 & 0.1878 & -\\
 ChatGLM2-FT  & 0.8576 & \underline{0.8141} & \underline{0.8352} & 0.2232 & 0.3589 & 0.2910 & \underline{$37.9\%$} \\
  \hline
 \textbf{Context-Aware (Ours)}  & \textbf{0.8628} & 0.8377 & 0.8505 & 0.2098 & 0.3502 & 0.2800 & $45.0\%$ \\
 \textbf{Intention-Enhanced (Ours)} & 0.8622 & 0.8390 & 0.8504 & 0.2041 & 0.3395 & 0.2718 & $\textbf{84.0\%}$ \\
 \multicolumn{1}{r|}{+ dynamic demo selection} & 0.8612 & \textbf{0.8527} & \textbf{0.8569} & 0.2105 & 0.3627 & 0.2866 & $82.0\%$ \\
 \hline
 Improvement (\%) & 0.07\% & 4.74\% & 2.60\% & - & - & - & $121.64\%$ \\ \hline
\end{tabular}}
\caption{Performance comparison of similar question generation methods. The universal best results are \textbf{bolded}, and the best results among baseline methods are \underline{underlined} to compute relative improvement.}
\label{table:experiment1}
\end{table*}

\section{Experimental Setup}
\label{sec:exp}

\subsection{Dataset}
\label{sec:data}

To evaluate the proposed methods for generating similar questions, we leverage a dataset sourced from an active customer service chatbot in the financial sector, which comprises over 3,000 QA pairs in Chinese, each with an average of 40 human-annotated similar questions. From this, we constructed a training dataset of 90,000 instances by randomly sampling the raw QA pairs, following the format outlined in Table \ref{table:prompt_template}. Additional experiments with public datasets are presented in Appendix \ref{sec:more results} for completeness and reproducibility.

\subsection{Evaluation Details}
\label{sec:evaluation}

For the quantitative evaluation, we utilized 90 unseen QA pairs, each with an average of 45 reference questions. In the human assessment, we collected 15 new questions from the recent records of the service chatbot, reflecting practical use cases. We report the following performance metrics:

\paragraph{Semantic Relevancy} Precision is the maximum BERTScore \cite{zhang2019bertscore} between each generated question and reference question, measuring semantic consistency. Recall is computed inversely, assessing semantic diversity. The F1 score measures the harmonic mean of precision and recall, balancing relevance and diversity.

\paragraph{Character-Level Diversity} We use Distinct-N \cite{li2015diversity} to evaluate lexical diversity and report the \textit{Distinct-1}, \textit{Distinct-2}, and their average, \textit{Distinct-Avg} score, counting unique N-grams in generated questions. Higher values indicate greater textual diversity.

\paragraph{Qualitative Evaluation} Five industry experts assess generated questions against source QA, marking acceptable ones based on the semantic consistency and syntactic diversity criteria\footnote{Metrics and evaluation criteria are detailed in Appendix \ref{sec:details of evaluation}. Implementation details are presented in Appendix \ref{sec:details}.}.

\section{Results and Discussions}
\label{sec:result}

\subsection{Main Results} 

Table \ref{table:experiment1} presents results for generating 20 similar questions. Most methods achieve high precision, with generated questions closely aligning with source semantics. However, baseline methods show low recall, indicating limited diversity as a key challenge in SQG. Static in-context learning methods underperform in both precision and recall due to irrelevant demonstrations. Fine-tuning with a one-to-one objective (ChatGLM2-FT) improves recall while maintaining precision, demonstrating the value of task-specific adaptation. The proposed one-to-many training objective (Context-Aware) enhances both precision and recall, and the Intention-Enhanced method further improves diversity and relevance. The inclusion of dynamic demonstration selection achieves state-of-the-art performance, surpassing zero-shot methods.

Human evaluation shows that general-purpose text-generation models, SimBERT and RoFormer-Sim, perform poorly, with only 20\% of generated questions meeting the acceptance criteria. ChatGLM2-FT improves this to 37.9\%, but still remains largely redundant and fails to meet practical needs. While the Context-Aware method excels in quantitative metrics, its impact on the acceptance ratio is modest. Introducing the customer's intention via the source answer significantly broadens the space of exploration, resulting in the largest number of usable similar questions. This emphasizes the importance of contextual information in improving relevance and diversity. 

Finally, character-level diversity shows that ChatGLM2-FT and the proposed methods outperform SimBERT, RoFormer-Sim, and ChatGLM2 (Few-Shot) in Distinct scores. Although the zero-shot ChatGLM2 achieves the highest Distinct score, it sacrifices consistency, as reflected in its low precision score due to the unconditioned generation process, which is not an ideal behavior.

\begin{figure}[]
  \centering
\includegraphics[width=0.95\columnwidth] {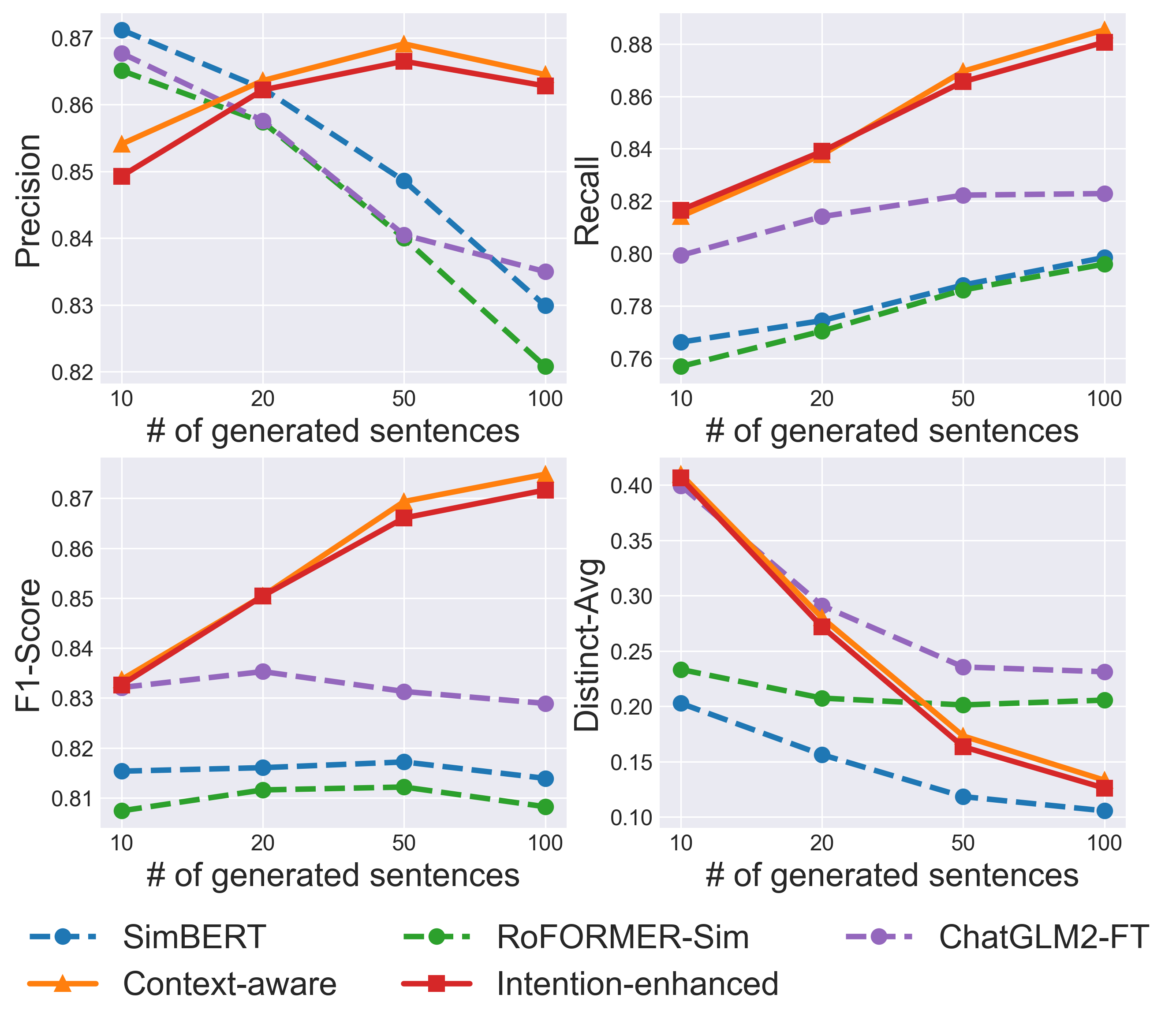}
  \caption{Performance comparison of similar question generation methods with varying number of questions.} 
  \label{fig:req_vq_number}
\end{figure}

\subsection{Performance vs. the Number of Generated Questions} 
\label{subsubsec: number}
In real applications, the desired quantity of similar questions often falls within the range of several tens to hundreds. Therefore, we systematically examine the performance of the proposed method by generating varying quantities of questions, up to a maximum of 100.

As shown in Figure \ref{fig:req_vq_number}, the Intention-Enhanced approach shows a surprising trend: precision stays consistently high, with only a slight decrease when generating up to 100 questions, while baseline methods experience a significant precision drop as the number of questions grows. This strength comes from the approach’s ability to balance relevance and variety, creating a diverse set of questions that closely match user intents and cover a broad range of query expressions. In contrast, baseline methods lose semantic consistency as question volume increases, reducing relevance and lowering precision. Additionally, our approach greatly improves recall, rising from $0.82$ to above $0.89$ as more questions are generated. Notably, our methods reach a recall of approximately $0.82$ when generating just 10 questions, surpassing baseline methods when generating 100 questions. This demonstrates that the one-to-many training objective effectively enables LLM to explore the semantic space surrounding the source question.


For character-level diversity, the proposed methods achieve the highest Distinct scores when generating 10 questions. All methods exhibit a reduction in distinct scores as the number of generated questions increases, which can be ascribed to the inherent constraint imposed by the limited length of the source question. For baseline methods, the declination is relatively modest, which can be attributed to the deviation of the semantic meaning. This observation also aligns with the decreased precision noted earlier and is further investigated through qualitative examples presented in Table \ref{table:example}.

\begin{table}[t]
\centering
\resizebox{\columnwidth}{!}{
\begin{tabular}{llcc}
\toprule
\textbf{Method} & \textbf{Constrain} & \textbf{Time} & \textbf{Total Diversity} \\
\midrule
Random & 20 choose 5 & 0:23 & 4.37\\
\textbf{Greedy (Ours)} & 20 choose 5 & 2:27 & 5.15\\ 
Exhaustive  & 20 choose 5 & 6:40:12 & 5.78\\
\hline
Random & 20 choose 10 & 1:39 & 20.14 \\
\textbf{Greedy (Ours)} & 20 choose 10 & 9:34 & 22.31 \\
Exhaustive & 20 choose 10 & - & - \\
\bottomrule
\end{tabular}}
\caption{Comparison of random selection and greedy algorithm in time efficiency and semantic diversity.}
\label{table:selection_methods}
\end{table}

\begin{CJK}{UTF8}{gbsn}
\begin{table*}[!t]
\footnotesize
\centering
\begin{tabular}{{p{2.5cm}|p{5.8cm}|p{5.8cm}}}
\toprule
  \textbf{Source question} & \multicolumn{2}{l}{证明开具时间要多久？ (How long does it take to process the certificate?)}  \\
  \\
 \textbf{Source Answer} & 
\multicolumn{2}{p{11.6cm}}{如您申请开具电子版证明，预计2个小时内发送至您指定的邮箱，纸质版证明开具时间预计3-8个工作日。 (If you apply for an electronic certificate, it is expected to be sent to your designated email address within 2 hours. The processing time for a hard copy certificate is estimated to be 3-8 working days.)}  \\
\midrule
\midrule
 \textbf{Method} & \textbf{High precision} & \textbf{Low precision} \\
\midrule
 \textbf{SimBERT} & 1. 证明要多久才可以开？ (How long does it take to obtain a certificate?) 2. 开证明一般要多久才能拿到？ (How long does it generally take to obtain a certificate?) 3. 一般证明需要多久才可以开？ (How long does it generally take to issue a certificate?) & 1. 证明要多长时间？(How long does it take for the proof?) 2. 公司证明怎么开？(How do I go about obtaining proof from the company?) 3. 证明书需要几个证明时间？ (How many proofing sessions are required for the proofing book?) \\
 & &\\
 \textbf{Context-Aware Batch Generation} & 1. 证明开具一般需要多长时间？ (How long does it typically take to obtain a certificate?) 2. 开具证明需要多久时间？(How long does it take to issue a certificate?) 3. 证明开具需要几日？ (How many days does it take to issue the certificate?) &   1. 我开了证明怎么还没收到？(I applied for the certificate, why haven't I received it yet?) 2. 当天可以开出证明吗？ (Can I get the certificate on the same day?) 3. 什么时候才能把证明发给我？ (When will I receive the certificate?) \\
 & &\\
 \textbf{Intention-Enhanced Batch Generation} & 1. 证明开具时间需要多久？ (How long does it take to process the certificate?) 2. 开具证明要多长时间？ (How long does it take to issue a certificate?) 3. 证明开具大概要多长时间？ (How long does it typically take to issue a certificate?) &  1. 今天开通证明，明天能发我吗？(If I request a certificate today, can it be delivered to me tomorrow?) 2. 今天可以开具电子证明吗？ (Can an electronic certificate be issued today?) 3. 开纸质证明要几天？(How many days does it take to issue a paper certificate?) \\
\bottomrule
\end{tabular}
\caption{Example demonstration of generated similar questions. For each method, we pick three questions with high precision (left) and three questions with low precision (right) to demonstrate semantic consistency and diversity.}
\label{table:example}
\end{table*}
\end{CJK}

\subsection{Selection of Generated Questions}
\label{sec:selection}

To select a subset of similar questions for knowledge base expansion, comparison in Table \ref{table:selection_methods}  highlights the superior diversity of the proposed greedy selection algorithm. Given that the augmentation is an offline process, we argue that the time invested in greedy search is a worthwhile trade-off for the improvement in total diversity. We also note that the exhaustive search method, which evaluates all possible subsets and guarantees optimality, is inefficient due to the NP-hardness of the problem, where selecting five questions from a set of 20 generated questions requires evaluating 15,504 combinations.

\subsection{Application Performance}
We evaluated the performance of our system in realistic customer service applications within the banking industry, deploying it for an in-app chatbot serving over 100,000 active users over a three-month period. The chatbot, equipped with a knowledge base augmented with the Intention-Enhanced approach, achieved the highest service success rate, with \textbf{over 92\% user satisfaction}, outperforming the unaugmented knowledge base at 74\% and the simple context-aware approach at 83.83\%. These results reveal the effectiveness of proposed augmentation strategies in enhancing service quality through improved query matching accuracy and more semantically consistent response retrieval.

\section{Qualitative Example Demonstration}

We demonstrate a typical example from customer service support to illustrate the effectiveness of the proposed methods, as shown in Table \ref{table:example}. The high-precision examples across all methods demonstrate strong semantic consistency with the source question, while the low-precision examples reveal notable distinctions. Questions generated by SimBERT deviate from the source question and exhibit a lack of fluency, exemplified by `How can I obtain a certificate from the company?' ( \begin{CJK}{UTF8}{gbsn}‘公司证明怎么开？’\end{CJK}). This departure from the original question is consistent with the significant drop in the recall score as seen earlier in Figure \ref{fig:req_vq_number}. Conversely, the Context-Aware Batch Generation method can generate novel expressions, such as `I applied for the certificate, why haven’t I received it yet?' ( \begin{CJK}{UTF8}{gbsn}‘我开了证明怎么还没收到？’\end{CJK}), which suggests a more effective exploration of the semantic space surrounding the source question. In the case of the Intention-Enhanced Batch Generation method, it becomes evident that information derived from the intention, not present in the source question, is effectively harnessed to generate similar questions, as shown by `electronic certificate' ( \begin{CJK}{UTF8}{gbsn}‘电子证明’\end{CJK}) and `paper certificate' (\begin{CJK}{UTF8}{gbsn}‘纸质证明’\end{CJK}). This underscores the importance of the customer intention as an effective guide to enhance the diversity of generated questions.

\section{Conclusion}
\label{sec:con}

This work presents an innovative approach to expanding the knowledge base of compliance-guaranteed service chatbots by generating similar questions with LLMs. By introducing a one-to-many training objective and utilizing customer intention as contextual guidance, we enhanced semantic diversity while staying aligned with the customer’s intent. The optimization framework enables our method to be seamlessly integrated into existing production systems, offering great flexibility and efficiency. These promising findings highlight LLMs' growing role in augmenting conventional system architectures in scenarios where standalone LLMs are not directly applicable, encouraging further research into LLM-guided systems for industrial deployment.

\section*{Limitations}
\label{sec:con}

While this work pioneers a novel strategy for augmenting a retrieval-based chatbot system with LLM-generated similar questions, it has two main limitations. First, it assumes customer inquiries are monolingual, overlooking the challenges of multilingual query matching, increasingly common in multinational enterprises and domains with frequent code-switching \cite{jiang2016cross}. Second, human evaluations by domain experts are costly and lack scalability. Recent studies suggest using LLM-as-a-judge to replace human involvement in performance evaluation, which could be better integrated into the proposed method to provide human-aligned feedback during question generation.

\section*{Acknowledgments}
This paper is partially supported by several ongoing projects led or coordinated by Prof. Zhang Chen, including P0048887 (Innovation and Technology Fund - ITSP, ITS/028/22FP), P0051906 (RGC Early Career Scheme, 25600624), and P0054482 (Two Square Capital Limited donation).

\bibliography{custom}

\newpage
\appendix

\section{Details on Evaluation Metrics}
\label{sec:details of evaluation}

For quantitative evaluation of the generated questions, we use the following metrics:

\begin{equation*}
precision = \sum_{i=1}^{n}{\frac{\text{max}_{j=1}^{m}{BERTScore(q_i,r_j)})}{n}}
\label{equ: precision}
\end{equation*}

\begin{equation*}
    recall= \sum_{i=1}^{m}{\frac{\text{max}_{j=1}^{m}{BERTScore(r_i,q_j)})}{m}}
\label{equ: recall}
\end{equation*}

\begin{equation*}
\label{equ: f1}
F1 = 2 \cdot \frac{precision \cdot recall}{precision + recall}
\end{equation*}

\begin{equation*}
\label{equ: distinct-n}
    Distinct\text{-}N = \frac{\text{count}(\text{unique}(N\text{-}grams))}{\text{count}(N\text{-}grams)}
\end{equation*}

\noindent For qualitative evaluation with human experts, the criteria can be summarized as: 1) \textbf{correctness}, ensuring that the generated questions are syntactically sound; 2) \textbf{relevancy}, which assesses whether the generated questions align with the original inquiry; and 3) \textbf{coherence}, which verifies whether the question should correspond to the same answer.

\section{Implementation Details}
\label{sec:details}

We utilize the lightweight ChatGLM2-6B~\cite{zeng2022glm} as our base model for its superior performance in the Chinese language \cite{hong2025qualbench}. Instead of employing parameter-efficient tuning~\cite{ding2022delta}, we implemented full tuning using Nvidia A100 GPUs on 90,000 training instances, as this approach yields significantly better performance. In the case of SimBERT and RoFormer-Sim, we set the temperature and top-k parameters to 0.9 and 5, respectively. For the ChatGLM2-based methods, we adhere to the default generation settings. The number of generated outputs, denoted as $L$, is set to 20, as we have observed that increasing this number often leads to degraded performance, which is discussed in Section \ref{subsubsec: number}.

\begin{table*}[!t]
\centering
\small
\begin{tabular}{lllcccc}
\toprule
\textbf{Domain} & \textbf{Model} & \textbf{Method} & \textbf{Precision} & \textbf{Recall} & \textbf{F1} & \textbf{Acceptance Ratio} \\
\midrule
\multirow{6}{*}{\textbf{telecom}} & \multirow{2}{*}{gpt-3.5} & context  & 0.6845 & 0.6908 & 0.6876 & 28\% \\
                                &                       & intent  & 0.6327 & 0.6492 & 0.6408 & 62\% \\
\cmidrule(lr){2-7}
                                & \multirow{2}{*}{gpt-4} & context & 0.7076 & 0.7157 & 0.7116 & 32\% \\
                                &                         & intent  & 0.6420 & 0.6526 & 0.6472 & 68\% \\
\cmidrule(lr){2-7}
                                & \multirow{2}{*}{chatglm} & context & 0.6050 & 0.5890 & 0.5969 & \textbf{44\%} \\
                                &                         & intent  & 0.5836 & 0.5708 & 0.5771 & \textbf{76\%} \\
\midrule[0.9px]
\multirow{6}{*}{\textbf{loan}} & \multirow{2}{*}{gpt-3.5} & context  & 0.7002 & 0.7645 & 0.7311 & 32\% \\
                                &                       & intent  & 0.6278 & 0.6890 & 0.6569 & 68\% \\
\cmidrule(lr){2-7}
                                & \multirow{2}{*}{gpt-4} & context  & 0.7223 & 0.7853 & 0.7526 & 38\% \\
                                &                         & intent  & 0.6370 & 0.6990 & 0.6663 & 72\% \\
\cmidrule(lr){2-7}
                                & \multirow{2}{*}{chatglm} & context  & 0.6314 & 0.6324 & 0.6319 & \textbf{46\%} \\
                                &                         & intent  & 0.5801 & 0.6302 & 0.6042 & \textbf{88\%} \\
\midrule[0.9px]
\multirow{6}{*}{\textbf{legal}} & \multirow{2}{*}{gpt-3.5} & context  & 0.6858 & 0.7416 & 0.7127 & 44\% \\
                           &                       & intent  & 0.6337 & 0.7048 & 0.6676 & 64\% \\
\cmidrule(lr){2-7}
                           & \multirow{2}{*}{gpt-4} & context  & 0.7035 & 0.7671 & 0.7341 & \textbf{62\%} \\
                           &                         & intent  & 0.6415 & 0.7187 & 0.6780 & \textbf{88\%} \\
\cmidrule(lr){2-7}
                           & \multirow{2}{*}{chatglm} & context & 0.7035 & 0.7671 & 0.7341 & 42\% \\
                           &                         & intent  & 0.6800 & 0.7401 & 0.7088 & 72\% \\
\bottomrule
\end{tabular}
\caption{Performance of different models across various domains for similar question generation.}
\label{table:model_performance}
\end{table*}

\section{More Results}
\label{sec:more results}
In this section, we present additional experimental results to gain further insights into industrial deployment and the general applicability of the proposed methods. The objective of these experiments is to demonstrate two crucial properties: 1) \textbf{Model invariance}, which ensures that the performance remains consistent regardless of the model type or size applied, and 2) \textbf{Domain invariance}, which ensures that the performance remains consistent across different domains or tasks. 

To explore these aspects, we incorporate two additional LLMs through the OpenAI API, GPT-3.5 and GPT-4, which are proprietary models with advanced capabilities. Given that these models are black-boxed, we implement the Intention-Enhanced prompting without model finetuning. For evaluation, we utilize three publicly available QA datasets covering distinct domains: telecommunications (telecom), banking services for loans (loan), and legal services (legal). For each source question, we generate 20 similar questions. 

The results presented in Table \ref{table:model_performance} demonstrate the varying performances of different models. GPT-4 generally outperforms GPT-3.5 due to its larger model size. However, the fine-tuned ChatGLM2 often exhibits competitive performance, especially with intent enhancement, attributed to task-specific fine-tuning. This highlights the importance of domain adaptation in achieving better text quality. However, larger LLMs show advantages when evaluated on generic domains, such as legal and law, benefiting from extensive pretraining data.

The generation speed of the methods varies considerably. Both proposed methods benefit from the one-to-many batch generation, resulting in significantly faster generation compared to the one-to-one approach. For instance, the average speed for Context-Based Batch Generation using GPT-3.5 is 5.83 seconds per item (i.e., source question), while Intention-Enhanced Batch Generation takes 6.35 seconds. In contrast, the same task using the one-to-one paradigm requires 50.23 seconds per item, significantly slower than the proposed methods. We do not present results for one-to-one methods due to their poor performance and slow generation.

Finally, we observe that the relative performance in terms of generation quality, speed, and retrieval capability remains consistent across different tasks and models. The larger models tend to produce better results, which is consistent with previous research; however, the improvements are not significantly large. Due to cost constraints, we still recommend a smaller fine-tuned model for accomplishing the Similar Question Generation tasks.

\section{Similar Question Selection Optimization}
\label{sec:optimization}

While increasing the number of similar questions can enhance the retrieval capabilities of a knowledge base, it also leads to significant redundancy. This redundancy not only inflates maintenance costs and increases storage requirements but also extends retrieval times and leads to unpleasant user experiences. This highlights the importance of selecting an optimal subset of similar questions that maximizes diversity while efficiently managing resource constraints. To address this, we propose an optimization framework that incorporates semantic relationships and a predefined budget constraint \( B \), which reflects the limitations of storage or retrieval power in practical applications.

\subsection{Problem Formulation}

The optimization framework is designed to maximize semantic diversity within a constrained resource budget \( B \), a key parameter representing the system's capacity to manage selected questions. This budget can be interpreted as either a \textbf{storage constraint}, limiting the number of questions based on their length (e.g., number of characters), or a \textbf{retrieval power constraint}, where computational resources or time available for retrieving answers are limited. As more similar questions are selected, both storage and retrieval complexity are elevated, leading to high maintenance costs and increased system latency due to the need to compute similarity measures.

Each candidate question \( q \in Q^{*} \) is associated with a cost \( \text{cost}(q) \), which quantifies its storage or retrieval requirement. To simplify the optimization, we normalize \( \text{cost}(q) = 1 \), allowing \( B \) to directly represent the maximum number of questions that can be selected. The selected subset of questions is denoted as \( S \), and the semantic distance between any two questions, \( q_a \) and \( q_b \), is represented by \( \text{dist}(q_a, q_b) \). The optimization problem is formally defined as:

\[
\begin{aligned}
& \max_{S \subseteq Q^{*}} \sum_{\substack{q_a, q_b \in S \\ q_a \neq q_b}} \text{dist}(q_a, q_b) \\
& \text{s.t.} \quad \sum_{q \in S} \text{cost}(q) \leq B.
\end{aligned}
\]

\subsection{Practical Implications and Advantages of Budget Constraints}

The introduction of \( B \) as a budget constraint provides a practical mechanism to balance diversity and resource efficiency in real-world systems. This parameter enables the framework to address key operational challenges while ensuring adaptability and robust performance across different scenarios.

\paragraph{Resource Efficiency} The budget \( B \) directly constrains the total cost of the selected questions, which can reflect storage or computational resources. This ensures that the solution remains feasible within the system's operational limits:
\begin{itemize}
    \item \textbf{Storage Efficiency:} In systems with limited storage capacity, \( B \) controls the total number of questions stored, prioritizing semantic diversity while minimizing redundancy. This leads to more efficient use of storage resources.
    \item \textbf{Retrieval Scalability:} By limiting the size of the selected subset, \( B \) reduces the computational complexity of pairwise similarity calculations during retrieval. This improves system responsiveness and ensures scalability for larger datasets.
\end{itemize}

\paragraph{Flexibility} The budget acts as a tunable parameter that can be adapted to specific application requirements. By adjusting \( B \), practitioners can fine-tune the trade-off between diversity, storage, and retrieval efficiency.

\subsection{Proof of Proposed Solution}
We first establish that the problem is NP-hard by reducing it from a well-known NP-hard problem. Next, we prove that the objective function is submodular, enabling the use of the proposed greedy algorithm as described in Algorithm \ref{alg:greedy}. Finally, we demonstrate the \( 1 - 1/e \) approximation bound of the Greedy Algorithm.
\subsubsection{NP-Hardness of the Problem}
\begin{theorem}
\label{core-function}
The problem of selecting a subset \( S \subseteq Q^{*} \) to maximize the sum of pairwise distances
\[
f(S) = \sum_{\substack{q_a, q_b \in S \\ q_a \neq q_b}} \text{dist}(q_a, q_b),
\]
subject to the budget constraint \( \sum_{q \in S} \text{cost}(q) \leq B \), is NP-hard.
\end{theorem}
\begin{proof}
We establish NP-hardness by reducing the problem from the Maximum Diversity Problem (MDP), a known NP-hard problem \cite{kuo1993analyzing}. The MDP involves selecting \( k \) elements from a set to maximize the sum of pairwise distances:
\[
f(S) = \sum_{\substack{q_a, q_b \in S \\ q_a \neq q_b}} \text{dist}(q_a, q_b), \quad |S| = k.
\]
In our problem, consider the special case where each element has uniform cost, i.e., \( \text{cost}(q) = 1 \) for all \( q \in Q^{*} \), and the budget constraint is \( B = k \). This simplifies to selecting exactly \( k \) elements from \( Q^{*} \), equivalent to the MDP. Since the MDP is NP-hard, and our problem generalizes it with arbitrary costs and a flexible budget constraint, our problem is at least as hard as the MDP. Moreover, the problem is in NP, as verifying a candidate solution \( S \) involves checking in polynomial time whether \( \sum_{q \in S} \text{cost}(q) \leq B \) and computing the total diversity \( \sum_{\substack{q_a, q_b \in S \\ q_a \neq q_b}} \text{dist}(q_a, q_b) \). Thus, the problem is NP-hard.
\end{proof}
\subsubsection{Submodularity of the Objective Function}
Although the problem is NP-hard, the objective function is submodular, a property that enables an efficient greedy algorithm to approximate the optimal solution.
\vspace{0.5em}
\begin{theorem}
The objective function
\[
f(S) = \sum_{\substack{q_a, q_b \in S \\ q_a \neq q_b}} \text{dist}(q_a, q_b)
\]
is submodular and non-decreasing.
\end{theorem}
\begin{definition}[Submodularity]
A set function \( f: 2^N \to \mathbb{R} \) is submodular if, for any \( A \subseteq B \subseteq N \) and any \( x \notin B \):
\[
f(A \cup \{x\}) - f(A) \geq f(B \cup \{x\}) - f(B).
\]
\end{definition}
\noindent This property reflects diminishing returns: the marginal gain from adding an element decreases as the set grows.
\begin{proof}
For \( f(S) \), the marginal gain from adding a new element \( x \) to a set \( S \) is:
\[
\Delta f(S, x) = \sum_{q \in S} \text{dist}(x, q).
\]
For \( A \subseteq B \subseteq Q^{*} \) and \( x \notin B \):
\[
\Delta f(A, x) = \sum_{q \in A} \text{dist}(x, q)\] 
\[\Delta f(B, x) = \sum_{q \in B} \text{dist}(x, q).
\]
Since \( A \subseteq B \), the terms in \( \Delta f(A, x) \) are a subset of those in \( \Delta f(B, x) \). Thus, the marginal gain from adding \( x \) decreases as the set grows, satisfying the submodularity condition:
\[
\Delta f(A, x) \geq \Delta f(B, x).
\]
Hence, \( f(S) \) is submodular. Additionally, \( f(S) \) is non-decreasing, as adding an element can only increase the sum of pairwise distances.
\end{proof}
\subsection{Complexity and Approximation Analysis}
The proposed greedy algorithm in Algorithm \ref{alg:greedy} efficiently selects a diverse subset under a budget constraint while achieving a provable approximation guarantee.
\subsubsection{Approximation Guarantee}
Given the submodularity and monotonicity of the objective function \( f(S) \), the greedy algorithm provides the following approximation bound:
\[
f(S_\text{greedy}) \geq \left(1 - \frac{1}{e}\right) f(S_\text{optimal}),
\]
where \( S_\text{greedy} \) is the solution from the algorithm, and \( S_\text{optimal} \) is the optimal subset. This ensures the algorithm achieves at least 63\% of the optimal solution.
\subsubsection{Complexity Analysis}
The computational complexity of the greedy algorithm is analyzed as follows. Computing the marginal gain for all \( n \) candidates in each iteration requires \( O(nk) \) operations, where \( k \) represents the size of the selected subset \( S \). The algorithm executes at most \( O(n) \) iterations, as each iteration selects one element. Consequently, the total complexity is \( O(n^2 k) \). With precomputed distances, the complexity reduces to \( O(n^2) \) at the cost of \( O(n^2) \) storage. The greedy algorithm balances solution quality and efficiency, providing near-optimal results with manageable computational overhead for moderate-sized datasets.
\end{document}